\documentclass[runningheads,a4paper]{llncs}

\usepackage{microtype}
\usepackage{amssymb}
\usepackage{graphicx}
\usepackage{tikz} 
\usepackage[small]{caption}
\usepackage[ruled]{algorithm2e}
\usepackage{booktabs}
\usepackage{amsmath}
\usepackage{mathabx}
\usepackage{dsfont}

\usepackage{url}
\urldef{\mailsa}\path|{tuennermann, enns,  mertsching}@get.upb.de|

\newcommand{\keywords}[1]{\par\addvspace\baselineskip
\noindent\keywordname\enspace\ignorespaces#1}

\newcommand*\circled[1]{\tikz[baseline=(char.base)]{
            \node[shape=circle,draw,inner sep=0.6pt] (char) {\small{#1}};}}

\newcommand*\circledsmall[1]{\tikz[baseline=(char.base)]{
            \node[shape=circle,draw,inner sep=0.6pt] (char) {\tiny{#1}};}}
            
\newcommand{\argmax}[1]{\underset{#1}{\operatorname{arg\,max}}}

\begin{document}

\mainmatter  

\title{Saliency-Guided Perceptual Grouping Using Motion Cues in Region-Based Artificial Visual Attention}
\titlerunning{Saliency-Guided Grouping in Region-Based Attention}

\authorrunning{T{\"u}nnermann et al.} 

%
%
\author{Jan T{\"u}nnermann \and Dieter Enns \and B{\"a}rbel Mertsching}
%



\institute{GET Lab, University of Paderborn, Pohlweg 47--49, 33098 Paderborn, Germany
\mailsa\\
\url{http://getwww.upb.de}
}

%
%

\toctitle{Lecture Notes in Computer Science}
\tocauthor{Authors' Instructions}
\maketitle

\begin{abstract}
Region-based artificial attention constitutes a framework for
bio-inspired attentional processes on an intermediate
abstraction level for the use in computer vision and mobile robotics.
Segmentation algorithms produce regions of coherently colored pixels.
These serve as proto-objects on which the attentional processes
determine image portions of relevance. A single region---which not
necessarily represents a full object---constitutes the focus of attention.
For many post-attentional tasks, however, such as identifying or
tracking objects, single segments are not
sufficient. Here, we present a saliency-guided approach that groups
regions that potentially belong to the same object based on proximity
and similarity of motion. We compare our results to object selection by
thresholding saliency maps and a further attention-guided strategy.
\end{abstract}

\section{Introduction}
Many artificial and natural systems must be able to visually select objects in dynamic scenes to perform cognitive tasks or actions. Selective visual attention, a concept from psychology, enables systems to distribute resources in a way that the relevant portions of a visual scene are processed efficiently. Stimuli that are unimportant in the current situation are disregarded or processed with low priority.

In biological systems, the focus of attention---the location in the visual field on which the available resources are focused---is determined with regard to bottom-up saliency and top-down influences. The latter are guided by knowledge with respect to the currenct task, whereas bottom-up saliency refers to local contrasts in the image that render certain locations conspicuous. Different feature dimensions contribute to bottom-up saliency, such as stimulus intensity or local orientation features. Activation from these different channels propagates towards a common saliency map; their integration is a prerequisite for the percept of a coherent object \nocite{Rensink2001,Treisman1980}.


Various approaches have been proposed to implement attention mechanisms for technical systems. The popular model by Itti et al. \cite{Itti1998} uses Difference-of-Gaussian and Gabor filters applied on image pyramids.  Local contrasts regarding the features color, intensity, and orientation (also motion and flicker in some versions, see e.g.\,\cite{Itti2004}) are computed by comparing pixels from finer scales to pixels of coarser scales. Weighting in the combination of the different feature dimensions can be used as a mean of top-down influence. That is, a representation of a search target can be learned as a set of weights for the combination which are then applied in the search (e,.g., see \cite{Kouchaki2012}). Other approaches for guiding attention in technical systems rely on frequency domain representations (see e.g., \cite{Cui2009}), use statistical methods (see e.g.,\,\nocite{Bruce2005}) or are region-based.


Region-based approaches perform an initial image segmentation to group similar pixels to coherent regions and then determine the focus of attention (FOA) based on theses regions. Different segmentation methods have been applied in different models, such as region growing \cite{Aziz2008bu}, super pixel methods \cite{Perazzi2012}, or colorspace quantization \cite{Cheng2011}. Some approaches allow the integration of top-down influences based template regions \cite{Aziz2008td,Tuennermann2013} or complexes of multiple regions \cite{Tuennermann2013td}.

An important feature is motion, as it indicates changes within dynamic scenes which may require the system to reorient. It therefore has been integrated in several attention systems applying pixel-based \cite{Belardinelli2009}, frequency domain \cite{Cui2009}, or region-based methods \cite{Tuennermann2012}  in a spatiotemporal context. The attention model by Tsotos et al. has been used to model attention towards motion in a biological plausible hierarchical manner \cite{Tsotsos2005}. This includes the progression from sensitivities for simple local translation to those for more complex patterns such as expanding or approaching stimuli. An elaborate survey on artificial attention systems and their biological foundations may be found in \cite{Frintrop2010}, whereas an extensive comparison of different technical systems was performed in \cite{Borji2012}.


Typically, the output of technical attention systems is a master saliency map that is a retinotopic mapping of the integrated activation from different feature channels and top-down influences. Additional modulation of this map may result from mechanisms such as inhibition of return (IOR) which suppresses locations (or features) that have been previously attended. The maximum activation at a certain time in the saliency map determines the FOA. Hence, the FOA is the point in the visual field that corresponds to the location of the saliency peak.

Many post-attentional tasks, however, require objects or their boundaries instead of a single point. For example, if the focus of attention shall be used to select an object and pass it to a typical object tracker, its boundaries are needed to obtain a suitable image patch. Also processes within the attention system benefit from more complex proto-objects, for example, when IOR is to be applied at the object level (to suppress attending of a whole object instead of an area that may or may not cover the object). The region-based approach is a step towards this, because instead of a single location the shape of the perceptual entity in focus is known. However, because these systems are based on segmentation with regard to some homogeneity criterion (e.g.\,similar color) and objects in general are not homogenous (e.g.\,textured objects), a region can capture full objects only in rare cases.

A common method to obtain objects after the attentional process is to select image portions that exceed a certain saliency threshold (see e.g., \cite{Cheng2011} or \cite{Perazzi2012}). This requires prior knowledge about what threshold will select a reasonable object. In cases where only a part of an object is salient, the thresholding procedure will select only this part and not the full object. To avoid such difficulties, Walther and Koch \cite{Walther2006} used a method that performs a segmentation at the FOA considering the low-level features which are already computed in the process of saliency computation. Their result is the approximate shape and extent of the object at the FOA.

For region-based attention, where segmentation of the scene already exits, no such method, which groups the pre-attentional segments to form an object at the FOA, is available yet. Only recently have multi-region objects been considered in region-based attention by \cite{Tuennermann2013td}; however, this method requires a given multi-region template which is then, on a sustained basis, attended throughout a sequence. A mechanism which is able to group regions around the FOA to form a coherent object independent from such a template is desirable for the mentioned reasons. 

In general, features of segments provide no unambiguous information about their relationship with regard to objects. In textured objects, features such as color, orientation, or symmetry carry little information about their membership in objects. Some features, however, such as a shared motion pattern or a common depth are highly indicative for regions of the same object, at least in a lot of situations. Thus, these features are well-suited to group individual regions to represent full objects.

In this paper, we extend the region-based spatiotemporal saliency model which was presented in \cite{Tuennermann2012}. The FOA obtained with this model is a single region which in most cases does not represent a full object. Here, we add a further step which, starting at the FOA, groups regions with similar motion to form coherent objects. 

Note that this concept differs from motion segmentation, which is a traditional topic in computer vision (see \cite{Zappella2008} for a review). In this line of research, the goal is to segment objects based on their motion for the whole scene. The concept we employ here, and which was also used by Walther and Koch \cite{Walther2006} whose approach is not limited to motion, is based on determining a relevant location first by means of saliency computation. An object representation is extracted only at this location. This is in line with studies of biological attention which show that attention is a prerequisite for representing objects (see e.g.\,\nocite{Rensink2001}).

\section{From Motion Saliency to Multi-region Objects}

\begin{figure*}
\centering
\includegraphics[width=1\textwidth]{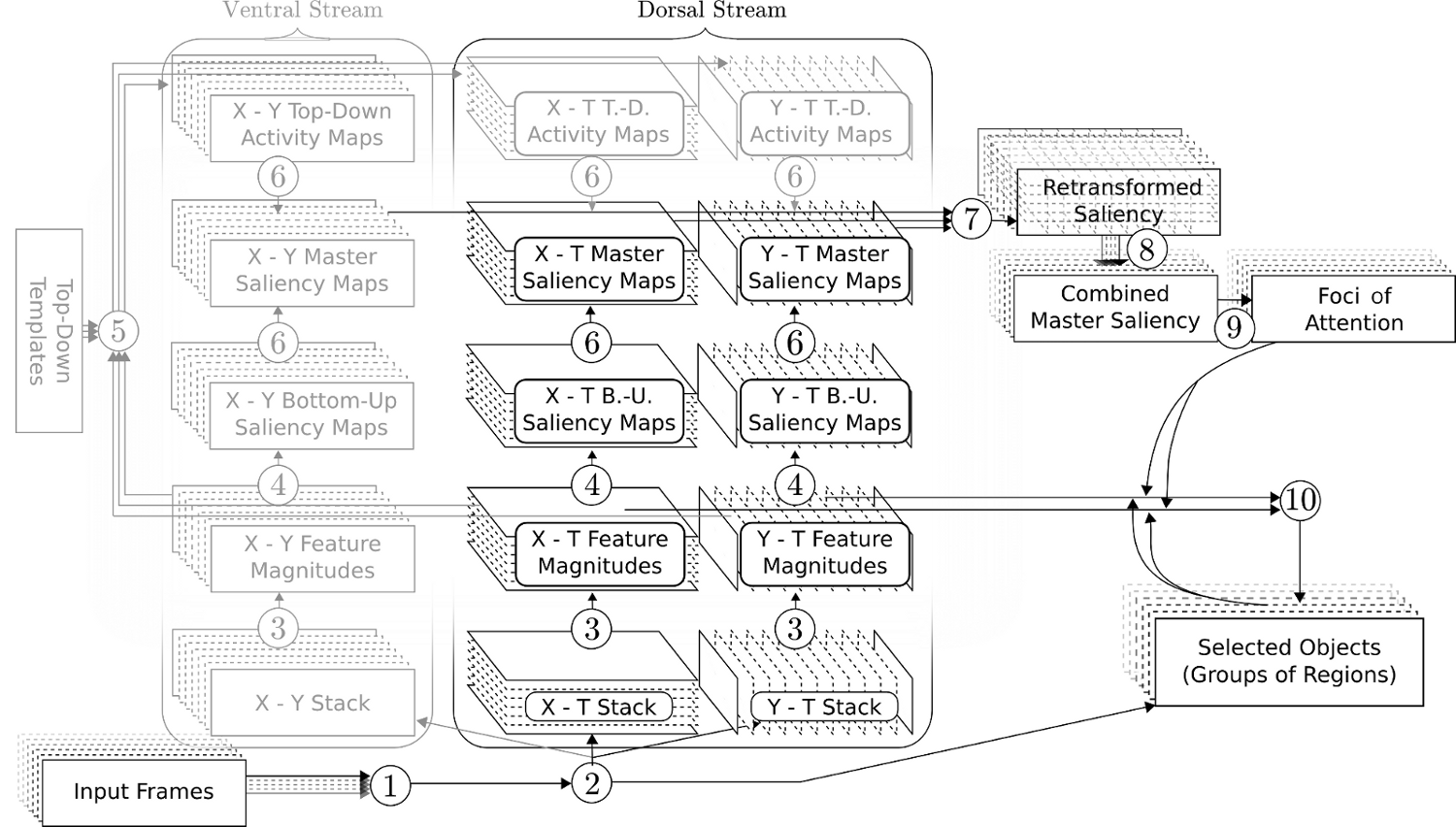}
\caption[]{A number of input frames forms a volume \circled{1} which is then sliced to produce $X-Y$,$X-T$, and $X-Y$ stacks which independently undergo a slice-wise color segmentation. Additionally, the $X-Y$ segmentation is forwarded to the object selection maps \circled{2}. The saliency processing is divided into a ventral and a dorsal stream, here only the latter one is used as it produces the motion saliency (see \cite{Tuennermann2013} for a description of the full two stream architecture). Feature magnitudes are calculated \circled{3} and based on these, feature and master saliency maps are produced (\circled{4} and \circled{6}). These maps are retransformed into the $X-Y$ image domain \circled{7} and combined \circled{8} to a stack of $X-T$ master saliency maps, one for each input frame from which the frames' FOAs are selected \circled{9}. Process \circledsmall{10} performs the grouping procedure. It selectively retrieves low-level feature magnitudes with the use of feedback from the FOA (about the 
initial location) and from the object selections maps (about neighboring regions which are candidates that are potentially added to the object representation). Parts of the architecture that have been grayed out refer to the dorsal stream which processes $X-Y$ feature saliencies (color, orientation, eccentricity, symmetry and size) or top-down influences. These mechanisms have not been used in the present study and we refer to \cite{Tuennermann2013} for their description.}
\label{archi}
\end{figure*}

\begin{figure}
\centering
\includegraphics[width=0.85\textwidth]{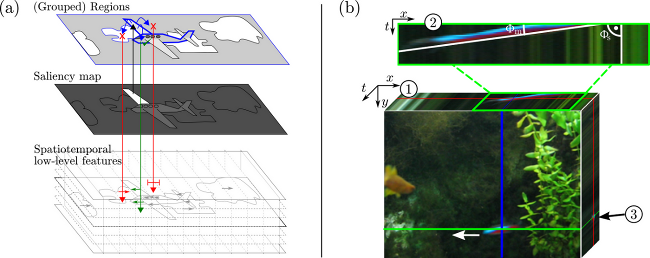}
\caption[]{ \textbf{(a)} Concept of selectively retrieving low-level features to build up a multi-region object representation at the FOA. The maximum saient region is fed forward (black arrow) to select a seed region. This is compared to its neighbors for which their spatiotemporal angle (reflecting their motion) is looked up. If it is similar to that of the seed region (green arrow), the procedure continues also for the neighbor (next iteration not shown). If the comparison fails (red arrows), the procedure terminates for these regions.  \textbf{(b)} The $X-Y-T$ cuboid \circled{1} shows slices from the inside of a frame volume on its faces. The white arrow indicates the direction of motion of the blue--red fish. The $X-T$ image \circled{2} shows a magnified excerpt of the $X-T$ slice. The white line drawn at the edge of the ``trace'' left by the fish has a spatiotemporal angle $\phi_m$ (with regard to the spatial axis) which is smaller than $90^\circ$ and corresponds to leftward motion. A line drawn along 
an 
edge produced by the static plant has an angle of $\phi_s = 90^\circ$, indicating that there is no motion. In the $Y-T$ slice, visualized on the right face of the cuboid, the fish appears as a small trace \circled{1} only, because it quickly passes through the slice.}
\label{concept}
\end{figure}

The concept for grouping regions at the FOA is shown in figure \ref{concept}: The saliency map contains one most salient region which is fed forward to select the region that represents the FOA. This region functions as a seed for the process that iteratively merges similarly moving regions. The spatiotemporal low-level features are looked up for the seed region and its immediate neighbors. If the absolute difference between the seed and a neighbor does not exceed a given threshold, the region is added to the group that represents the object and becomes a seed for a subsequent step. If for a neighbor the difference exceeds the threshold, the merging is stopped there. In the following, we formally describe the calculation of the motion feature as well as the saliency calculation and the merging process. The general data flow in the system is outlined in figure \ref{archi}.

\subsection{Motion Feature Magnitudes}\label{motion_feat}
The motion feature is calculated as described in \cite{Tuennermann2012} by applying the methods of region-based attention \cite{Aziz2008bu} on spatiotemporal slices. Spatiotemporal slices ($X-T$ and $Y-T$) are extracted from a pixel volume ($X-Y-T$) which is obtained by stacking $T$ ($10$ in our implementation) input frames ($X-Y$, $320\times240$ in our experiments). After this step, three stacks of 2-dimensional images are available: the $X-Y$ stack with $T$ slices which contains the original input frames, the $X-T$ stack with $Y$ images, and the $Y-T$ stacks with $X$ images. The $X-T$ and $Y-T$ stacks encode spatiotemporal behavior of objects, such as the horizontal and vertical components of motion, respectively. An exemplary $X-Y-T$ volume, spatiotemporal slices, and how moving objects appear in them are shown in figure \ref{concept}. For continuous input, such volumes are formed and processed one after the other.

Region lists $\Re$ with regions $R_i (i \in \{1 \dots n\})$ are created for each slice by applying a color segmentation procedure which groups similarly colored pixels to form regions (please refer to \cite{Aziz2005} or \cite{Backer2013} for details regarding such segmentation methods which have been used in region-based attention). 
Additionally, a label map $\mathds{L}$ that maps a position in the slice $(n,m)$ to a region $R_i$ is generated.

The motion feature is based on the concept of spatiotemporal receptive fields that respond to motion. The angle of an edge (or in our case a region) on a spatiotemporal slice is related to the motion of the corresponding object (see \cite{Adelson,Belardinelli2009} for details regarding this concept). 

\noindent Thus, we determine the orientation for each region on a spatiotemporal slice. This is done by determining the first (lowest t-coordinate) and last (highest t-coordinate) row of pixels from each region. In these rows the center pixels $c^{first}_i$ and $c^{last}_i$ are located and the region's spatiotemporal angle  $\phi^{st}_i$  is obtained as

\begin{equation}
 \displaystyle \phi^{st}_i = \text{atan2}( h , c^{first}_i - c^{last}_i )	
\end{equation}

\noindent where $\text{atan2}(t,x)$ calculates $\arctan{(t  x^{-1})}$ and adjusts the result to give the angle between $(x,t)$ and the positive $x$ axis. If $\phi^{st}_i$ is close to $90^\circ$, the corresponding object did not move with regard to the motion component represented in the slice ($X-T$ or $Y-T$).  Motion towards the left (or upwards, respectively) produces angles between $0^\circ$ and $90^\circ$, while motion towards the right (or downwards, respectively) results in values between $90^\circ$ and $180^\circ$.

This spatiotemporal feature is used to calculate the region-based saliency on each spatiotemporal slice as described in the next section. Eventually the spatiotemporal saliency is brought back into the $X - Y$ image domain to select the FOA.

\subsection{Motion Saliency}
The bottom-up motion saliency $\mathord{\uparrow}\mbox{MS}_i$ is computed as the sum of the normalized differences of the angles associated with region $i$ and every other region, that is,

\begin{equation}
  \displaystyle \mathord{\uparrow}\mbox{MS}_i = \sum_{j=1}^{\vert R \vert} {{\vert\phi^{st}_i - \phi^{st}_j\vert\over{180^\circ} } w^\Delta_{ij}}
\end{equation}
where $w^\Delta_{ij}$  ($0$ \dots $1$) is proportional to the distance between the centers of $R_i$ and $R_j$ and thus models that closer neighbors contribute more to a region's saliency. This motion saliency is obtained independently for the $X -T$ and $Y - T$ stacks and is then transformed back into the $X -Y$ domain by applying algorithm \ref{algo1}.

 \begin{algorithm}
\ForEach{region $R^{{xy}_t}_i \in \Re^{{xy}_t}$ ($t^{th}$ $X-Y$ frame)}
{
	\For{posY = $top^{{xy}_t}_i$ \textbf{to} $bottom^{{xy}_t}_i$} 
	{
		\For{posX = $left^{{xy}_t}_i$ \textbf{to} $right^{{xy}_t}_i$}  
		{
			\If{getLabel($\mbox{labels}^{xy}_t$,\,posX,\,posY) == $i$} 
			{
			 $\mbox{index}^{xt}$  = getLabel($\mbox{labels}^{xt}_y$,\,posX,\,t)\\
			 $\mbox{index}^{yt}$ = getLabel($\mbox{labels}^{yt}_x$,\,posY,\,t)\\
			 $\mathord{\uparrow}\mbox{MS}^{{xy}_t}_i += (\mathord{\uparrow}\mbox{MS}^{{xt}_\text{posY}}_{\text{index}^{xt}} +
			 \mathord{\uparrow}\mbox{MS}^{{yt}_\text{posX}}_{\text{index}^{yt}}) \cdot \frac{1}{2} \cdot \frac{1}{size^{{xy}_t}_i}$ 
			}
		}
	}
}
\caption{\footnotesize  This algorithm establishes correspondences between spatiotemporal saliency from $X-T$ and $Y-T$ slices and the $X-Y$ frames. The function \textit{getLabel($l$,$m$,$n$)} returns the label stored in label map $l$ at position ($m$,$n$) and mapping a position to the index (label) of the region in the corresponding region list.  Variables $left^{{xy}_t}_i$,$top^{{xy}_t}_i$, $right^{{xy}_t}_i$ and $bottom^{{xy}_t}_i$ refer to the bounding box of the region. Variable $size^{{xy}_t}_i$ contains the size of the region in pixels. This algorithm may be modified to also include saliency regarding spatial features (such as color, orientation, etc.; see \cite{Tuennermann2013} for details).}\label{algo1}
\end{algorithm}

The algorithm loops over all pixels within the bounding box of a each region (which is obtained during the segmentation process). A check is performed if the pixel belongs to the current region based on the label map. If this check is positive, the spatiotemporal saliencies are looked up in the $X -T$ and $Y - T$ stacks at the corresponding positions. These values are summed up for the region and the result is normalized by the region size.

The result is a motion saliency value $\mathord{\uparrow}\mbox{MS}^{{xy}_t}_i$ for every region that includes $X-T$ and $Y-T$ contributions. Here, motion is the only contributing feature. However, spatial features such as color, orientation, and size, as well as top-down influences in the form of search targets or feature weighting (see \cite{Tuennermann2013}) can also be integrated.

The FOA for frame $t$ is obtained as the region with the maximum overall saliency $R_m^{{xy}_t}$, with $\argmax{m}$ $\mathord{\uparrow}\mbox{MS}^{{xy}_t}_{m}$.
 
\subsection{Grouping Regions at the FOA}\label{proposed}
In order to group similarly moving regions at the FOA to form a coherent object, we apply a strategy that starts with the most salient region as a seed and selects its immediate neighbors (this relationship is established during the segmentation) as candidates which are added to a list of open regions $Q$. These are tested and if they fulfill a set of conditions, they are added to the list $O$ that represents the object. Their neighbors are added to the list $Q$. Once a region was tested, it is removed from $Q$ and the procedure terminates when $Q$ is empty. 

The conditions to be fulfilled for a region $i$ to be added to the object representation $O$ are given in the following list:
\begin{description}
 \item [(a)] $\sqrt{({\mathord{\leftrightarrow}\varphi}^{{xy}_t}_i)^2 + ({\mathord{\updownarrow}\varphi}^{{xy}_t}_i)^2} < \tau$. Where $({\mathord{\leftrightarrow}\varphi}^{{xy}_t}_i)$ and $({\mathord{\updownarrow}\varphi}^{{xy}_t}_i)$  are the averaged spatiotemporal angles which are obtained by looping over all pixels of a region and summing and normalizing the $X-T$ and $Y-T$ contributions as shown in algorithm \ref{algo2}. Threshold $\tau$ is set to $44$ in our implementation.
 \item [(b)] $\left|{(\mathord{\leftrightarrow}\varphi}^{{xy}_t}_i)-90^\circ \right| > \sigma_{xt}$ and $\left|{(\mathord{\updownarrow}\varphi}^{{xy}_t}_i)-90^\circ\right| > \sigma_{yt}$. Thus, the $\sigma$ thresholds enforce a minimal difference to the spatiotemporal angle of $90^\circ$ (no motion) to exclude contributions from noise. We use $\sigma_{xt}=\sigma_{yt}=10^\circ$.  
 \item [(c)] The size of the group of a candidate $i$ merged with all regions currently in $O$ (current size) must be less than $\eta$ times the current maximum size that has been discovered for $O$ on previous frames (this check is not performed for the first frame of each volume). In our tests we used $\eta=1.5$. 
 \end{description}

\begin{algorithm}
\For{posY = $top^{{xy}_t}_i$ \textbf{to} $bottom^{{xy}_t}_i$} 
{
	\For{posX = $left^{{xy}_t}_i$ \textbf{to} $right^{{xy}_t}_i$}  
	{
		\If{getLabel($\mbox{labels}^{xy}_t$,\,posX,\,posY) == $i$} 
		{
		  $\mbox{index}^{xt}$  = getLabel($\mbox{labels}^{xt}_y$,\,posX,\,t)\\
		  ${\mathord{\leftrightarrow}\varphi}^{{xy}_t}_i  += \phi^{{xt}_\text{posY}}_{\text{index}^{xt}} \cdot \frac{1}{size^{{xy}_t}_i }$ \\ 
		  $\mbox{index}^{yt}$ = getLabel($\mbox{labels}^{yt}_x$,\,posY,\,t)\\
		  ${\mathord{\updownarrow}\varphi}^{{xy}_t}_i  += \phi^{{yt}_\text{posX}}_{\text{index}^{yt}} \cdot \frac{1}{size^{{xy}_t}_i}$  
		
		}
	}
}
\caption{\footnotesize For a $X-Y$ region $i$, this algorithm performs a look-up of the low level motion signatures. These are summed over all the spatiotemporal locations that correspond to the position of $i$ in frame $t$. The function \textit{getLabel($l$,$m$,$n$)} returns the label stored in label map $l$ at position ($m$,$n$) and mapping a position to the index (label) of the region in the corresponding region list. Variables $left^{{xy}_t}_i$,$top^{{xy}_t}_i$, $right^{{xy}_t}_i$ and $bottom^{{xy}_t}_i$ refer to the bounding box of the region. Variable $size^{{xy}_t}_i$ contains the size of the region in pixels.}\label{algo2}
\end{algorithm}

\vspace{-0.6cm}
\noindent Thus, besides proximity, which is enforced by the neighborhood relation, condition (a) represents the check for having a similar motion signature in the spatiotemporal domain. Condition (b) is the requirement for a minimum deviation of the spatiotemporal angle from that of a perfect static object. The $\sigma$ parameters can be regarded as thresholds below which the magnitudes of the spatiotemporal angles are considered not reliable. Condition (c) is a sanity check that ensures that an object cannot dramatically increase its size from one instance to the next. Note that this check assumes that the FOA focuses on the same object for the whole volume. Given the fact that we use short volumes of ten frames, this is usually the case.

\begin{figure*}
\centering
\includegraphics[width=0.9\textwidth]{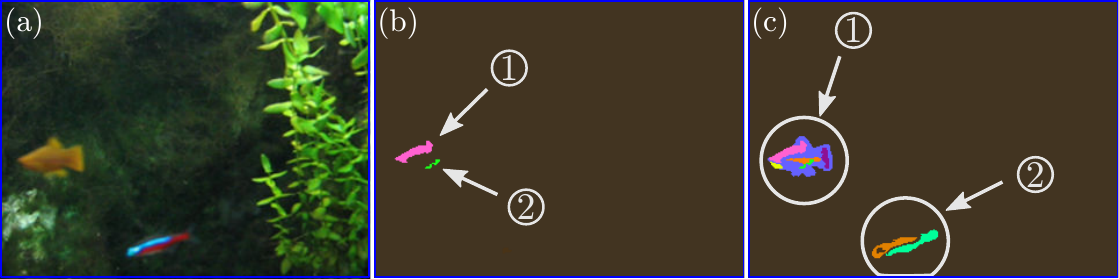}
\caption[]{ \textbf{(a)} Input frame.  \textbf{(b)} and  \textbf{(c)} Maps showing selected regions in random colors after the first cycle \circled{1} and second cycle \circled{2} with  \textbf{(b)} region-based IOR and  \textbf{(c)} object-based IOR. The white annotations have been manually added.}\label{IOR}
\end{figure*}

Depending on the application, a frame-based inhibition of return (IOR) may be useful to attend further objects within a frame (as we demonstrate in section \ref{test2}). To achieve this, after the first object at the FOA is selected, the saliency of the corresponding regions is tuned down and the procedure is repeated to select the next object on the current frame. Such an object-based IOR is highly advantageous compared to a region-based IOR. This can be seen in figure \ref{IOR}c where the second cycle discovers a new object while in figure \ref{IOR}b only another part of the same object is revealed. Note that inhibiting object selection over a range of frames requires location-based inhibition which could be implemented by selecting and inhibiting regions and their projected future locations (using the present information about their motion). Such a mechanism is not implemented in the current version. 

An important aspect of the proposed concept is that correspondence between a perceptual entity (region) and a low-level feature (here spatiotemporal angle, i.e., motion) is established ``on demand'' starting at the FOA. Other features, such as depth, may be integrated in the same manner, restricting their computation to the locations which have proven to be salient. 

\vspace{-0.5cm}
\section{Experimental Results}
In the tests described in the following sections, we run our algorithm with the parameters as described in section \ref{proposed}. The region-growing segmentation procedure which is used as part of the computation of the motion feature and the motion saliency is controlled by a set of thresholds. These define the allowed color difference between seed and candidates as well as the current border and candidates. The concrete values of these thresholds depend on the hue--saturation--intensity values found in the images (e.g.\,different thresholds at different hues). However, the parameter set was kept constant for all tests.
{\sloppy
We compare our results with object selections by thresholding saliency maps from the motion saliency approach \cite{Tuennermann2012} on  which the FOA determination in the proposed system is based.} Furthermore, we use saliency maps generated by the model by Itti et al. \cite{Itti2004,Itti1998}\footnote{\scriptsize The \texttt{ezvision} software [from \url{http://ilab.usc.edu/toolkit}, June, 2013] was used to obtain results for the saliency model from Itti et al. \cite{Itti2004,Itti1998} and Walther and Koch \cite{Walther2006}.}. Additionally to the default version which uses color-, flicker-, intensity-, orientation-, and motion-channels (CFIOM), we generated saliency maps using only the motion channel (M), because we are dealing only with moving objects.

In the thresholding procedure, a value is incremented and locations in the saliency maps with higher values are considered to belong to the object. This is then compared with ground truth from frames in which the moving foreground objects have been manually marked. The true positive rate is plotted against the false positive rate for the different saliency models providing a rough context for judging the performance of the proposed algorithm and the approach from Walther and Koch \cite{Walther2006} that also aims at selecting objects at the FOA. 

\subsection{Test 1: One Moving Object}

\begin{figure*}
\centering
\includegraphics[width=1\textwidth]{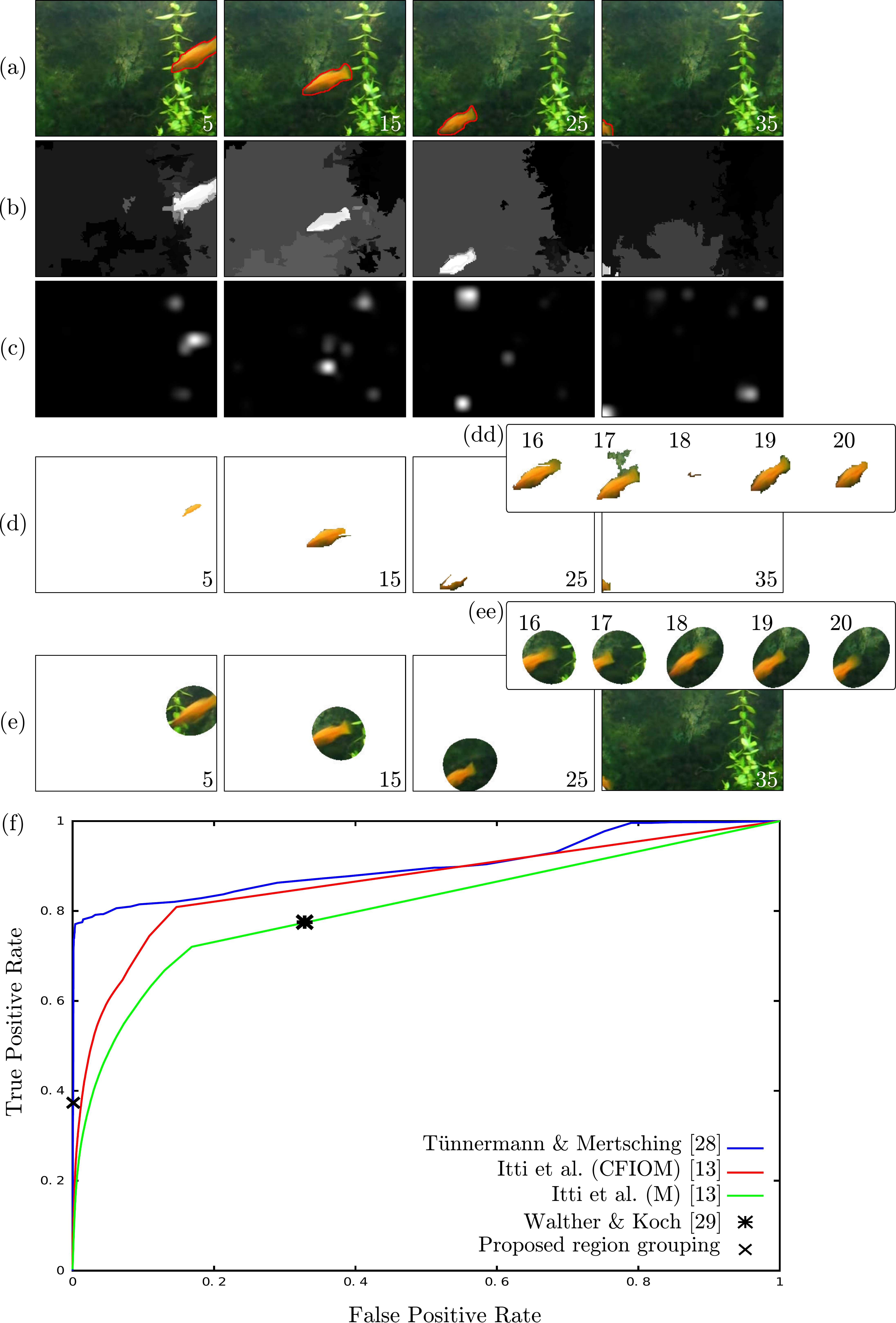}
\caption[]{ \textbf{(a)} Exemplary input images of a sequence with ground truth marked (true positives are inside the red boundary).  \textbf{(b)} Corresponding saliency maps of \cite{Tuennermann2012} (normalized for visualization).  \textbf{(c)} Saliency maps of Itti et al. \cite{Itti2004,Itti1998} with all channels (CFIOM).  \textbf{(d)} Objects selected by the proposed algorithm at their location.  \textbf{(dd)} Isolated selection on continuous frames in the middle of the scene.  \textbf{(e)} Objects selected by the algorithm of Walther and Koch \cite{Walther2006}.  \textbf{(ee)} Isolated selections on continuous frames in the middle of the scene.  \textbf{(f)} Performances of the object selection approaches compared to thresholding saliency maps of different algorithms.}\label{one}
\label{scene1}
\end{figure*}

In the first scene ($40$ frames), there is a single moving foreground object (a fish) which is reliably attended by the  region-based motion saliency approach. We refer to \cite{Tuennermann2013} where this scene has also been processed with the motion saliency procedure and the results show that only small parts of the object are selected. In figure \ref{one}d and \ref{one}dd  results from the proposed procedure are shown; results from Walther and Koch \cite{Walther2006} are shown in \ref{one}e and \ref{one}ee. The input and ground truth is shown in figure \ref{one}a and saliency maps from \cite{Tuennermann2012} and \cite{Itti2004} in figure \ref{one}b and \ref{one}c, respectively.

It can be seen that the proposed approach sometimes captures the entire object, but in general, it is rather conservative which is also reflected in a false positive rate close to zero (see figure \ref{one}f). The algorithm from \cite{Walther2006} selects the full target more often (higher true positive rate), but also selects a substantial amount of background which is reflected in a larger false positive rate. A reason for the rough and frequently too large shape boundaries of this method might be the low resolution of the internal feature maps used by the model.

\noindent Figure \ref{one}f also shows that the object selection methods in average perform similarly as their related saliency approaches at certain thresholds. Still, the  object selection approaches are advantageous in practice when the optimal threshold is unknown or must be chosen conservatively.

\subsection{Test 2: Two Moving Objects}\label{test2}
\begin{figure*}
\centering
\includegraphics[width=.9\textwidth]{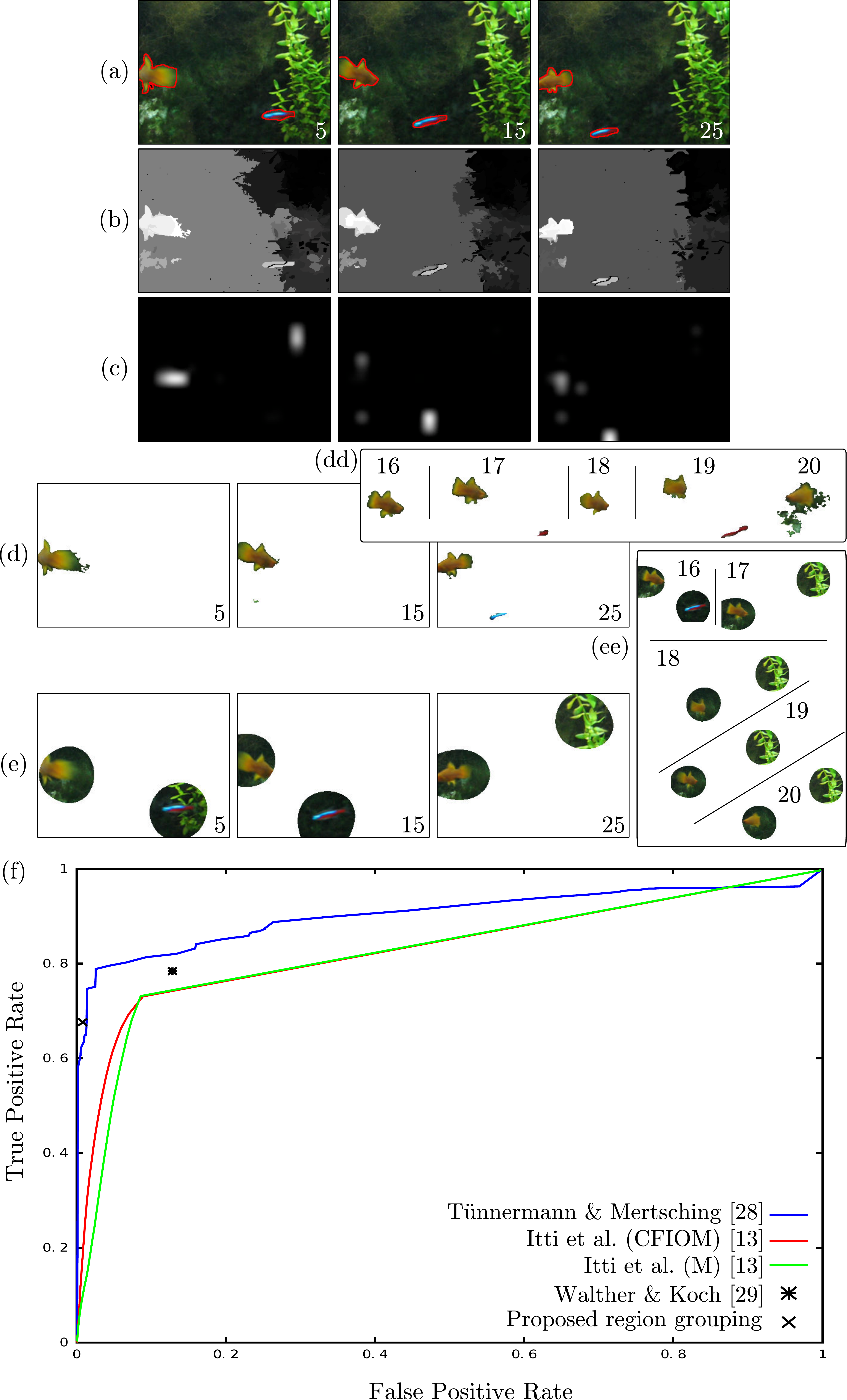}
\caption[]{ \textbf{(a)} Exemplary input images of a sequence with ground truth marked (true positives are inside the red boundary). {b} Corresponding saliency maps of \cite{Tuennermann2012} (normalized for visualization).  \textbf{(c)} Saliency maps of Itti et al. \cite{Itti2004,Itti1998} with all channels (CFIOM).  \textbf{(d)} Objects selected by the proposed algorithm at their location.  \textbf{(dd)} Isolated selection on continuous frames in the middle of the scene.  \textbf{(e)} Objects selected by the algorithm of Walther and Koch \cite{Walther2006}.  \textbf{(ee)} Isolated selections on continuous frames in the middle of the scene.  \textbf{(f)} Performances of the object selection approaches compared to thresholding saliency maps of different algorithms.}\label{two}
\end{figure*}

The second test scene ($30$ frames) features two moving foreground object (two fish). All parameters were kept the same as in the previous test except that the proposed algorithm is allowed a second cycle per frame to make use of the object-based IOR (see figure \ref{IOR}). The model of Walther and Koch \cite{Walther2006} was also applied two times for each frame. The timing of the attention system was adjusted so that---with the help of shape-based IOR (see  \cite{Walther2006})---two objects were selected on each frame. Again it can be seen (selections shown in figure \ref{two}d, \ref{two}dd, \ref{two}e, and \ref{two}ee; performance plot \ref{two}f) that the proposed region-based method is more likely to miss parts of the target whereas the method by \cite{Walther2006} returns the targets merged with some surrounding background. It also sometimes selects the plant which is no moving foreground object, probably because their method is not limited to objects in motion and the plant is highly contrasting in 
intensity. 

\section{Conclusion and Outlook}
\vspace{-0.4cm}
A conclusion drawn from our tests is that the region-based grouping strategy, which considers shared motion patterns, yields rather conservative results. In some frames (see figure \ref{two}ee) objects are selected in close agreement with their actual shape. The evaluated scenes were rather simple, containg no ego motion of the system. An additional test was performed with an object which is static with respect to the image in front of a dynamic background (relative motion; the same scene can be seen in figure 10 in \cite{Tuennermann2013}). The test revealed that such situations are rather difficult for the current implementation because the background regions, which are crossed by the target, show noisy spatiotemporal angles (note that condition (a) described in section \ref{proposed} had to be ignored, because the target was static). The model performed with 17 \% true positives and no false positives. However, the model by Walther and Koch \cite{Walther2006} also showed a weak performance for this 
scenario (26 \% TP and 23 \% FP). For future improvement of our system, we are investigating the optimization of segmentation methods for spatiotemporal slices and more stable calculations of the spatiotemporal angles.

\begin{figure*}
\centering
\includegraphics[width=.8\textwidth]{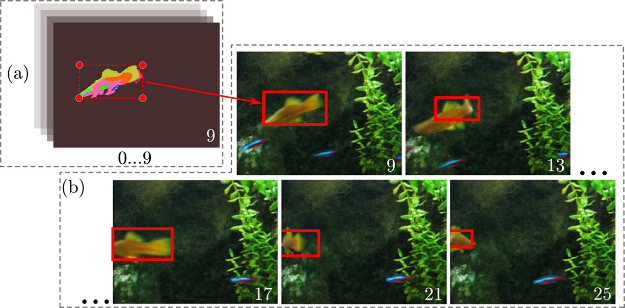}
\caption[]{ \textbf{(a)} The bounding box is determined for a group of regions generated with the proposed algorithm. \textbf{(b)} The bounding box is used to select (red arrow) an image patch for the TLD \nocite{Kalal2010} tracking procedure which is applied on subsequent frames. The fish is successfully tracked also in the frames between those which are shown.}\label{TLD}
\end{figure*}

In figure \ref{TLD}, we show results of an object tracking \nocite{Kalal2010} that has been initialized with such a selection from the proposed system. A direction for further future work is to determine the reliability of a selection and only feed good representations to higher-level processes such as object detection or tracking.
\vspace{-0.3cm}

\vspace{-0.3cm}
\noindent Further improvements could be gained by including additional features. In agreement with the proposed concept, depth could be estimated locally (e.g., by determining local stereo correspondences) starting at the FOA.

\bibliographystyle{splncs}
\bibliography{bibliography}

\end{document}